\documentclass[final]{cvpr}

\usepackage{times}
\usepackage{epsfig}
\usepackage{graphicx}
\usepackage{amsmath}
\usepackage{amssymb}
\usepackage{comment}

\usepackage{multirow}
\usepackage{makecell}
\usepackage{adjustbox}
\usepackage{bbding}
\usepackage{nopageno} 

\usepackage[pagebackref=true,breaklinks=true,colorlinks,bookmarks=false]{hyperref}

\def\cvprPaperID{2564} 
\def\confYear{CVPR 2021}

\newcommand{\best}[1]{\color{red}\textbf{#1}}
\newcommand{\second}[1]{\color{blue}\textbf{#1}}

\begin{document}
\title{Generalized Few-Shot Object Detection without Forgetting}

\author{{Zhibo Fan},
        {Yuchen Ma}, \ 
        {Zeming Li}, 
        {Jian Sun} \\
        
        \vspace{4pt}
        Megvii Technology \\
       \vspace{4pt}
       \small{zb1439@outlook.com,\{mayuchen, lizeming, sunjian\}@megvii.com}}

\maketitle

\begin{abstract}
   Recently few-shot object detection is widely adopted to deal with data-limited situations. While most previous works merely focus on the performance on few-shot categories, we claim that detecting all classes is crucial as test samples may contain any instances in realistic applications, which requires the few-shot detector to learn new concepts without forgetting. Through analysis on transfer learning based methods, some neglected but beneficial properties are utilized to design a simple yet effective few-shot detector, Retentive R-CNN. It consists of Bias-Balanced RPN to debias the pretrained RPN and Re-detector to find few-shot class objects without forgetting previous knowledge. Extensive experiments on few-shot detection benchmarks show that Retentive R-CNN significantly outperforms state-of-the-art methods on overall performance among all settings as it can achieve competitive results on few-shot classes and \textbf{does not degrade the base class performance at all}. Our approach has demonstrated that the long desired never-forgetting learner is available in object detection.
\end{abstract}


\vspace{-10pt}
\section{Introduction}
Computer vision community has seen significant progress by applying deep convolutional neural networks trained from a massive amount of data. 
However, sufficient training data is sometimes unavailable due to extensive human labor for annotation, especially for object detection, and the source data distribution may be long-tailed by nature such that certain object categories only contain limited examples.
These circumstances raise the need to learn under a low-data regime effectively. Inspired by human's ability to learn new concepts rapidly from a handful of examples, few-shot learning\cite{fei2006one,koch2015siamese,vinyals2016matching,snell2017prototypical,sung2018learning,qi2018low,gidaris2018dynamic,finn2017model,rusu2019meta,dhillon2019baseline,chen2020new,wang2019simpleshot} is then proposed to mimic such generalization capability, with extensive research on image classification.

\begin{figure}[t]
\begin{center}
\includegraphics[width=1\linewidth]{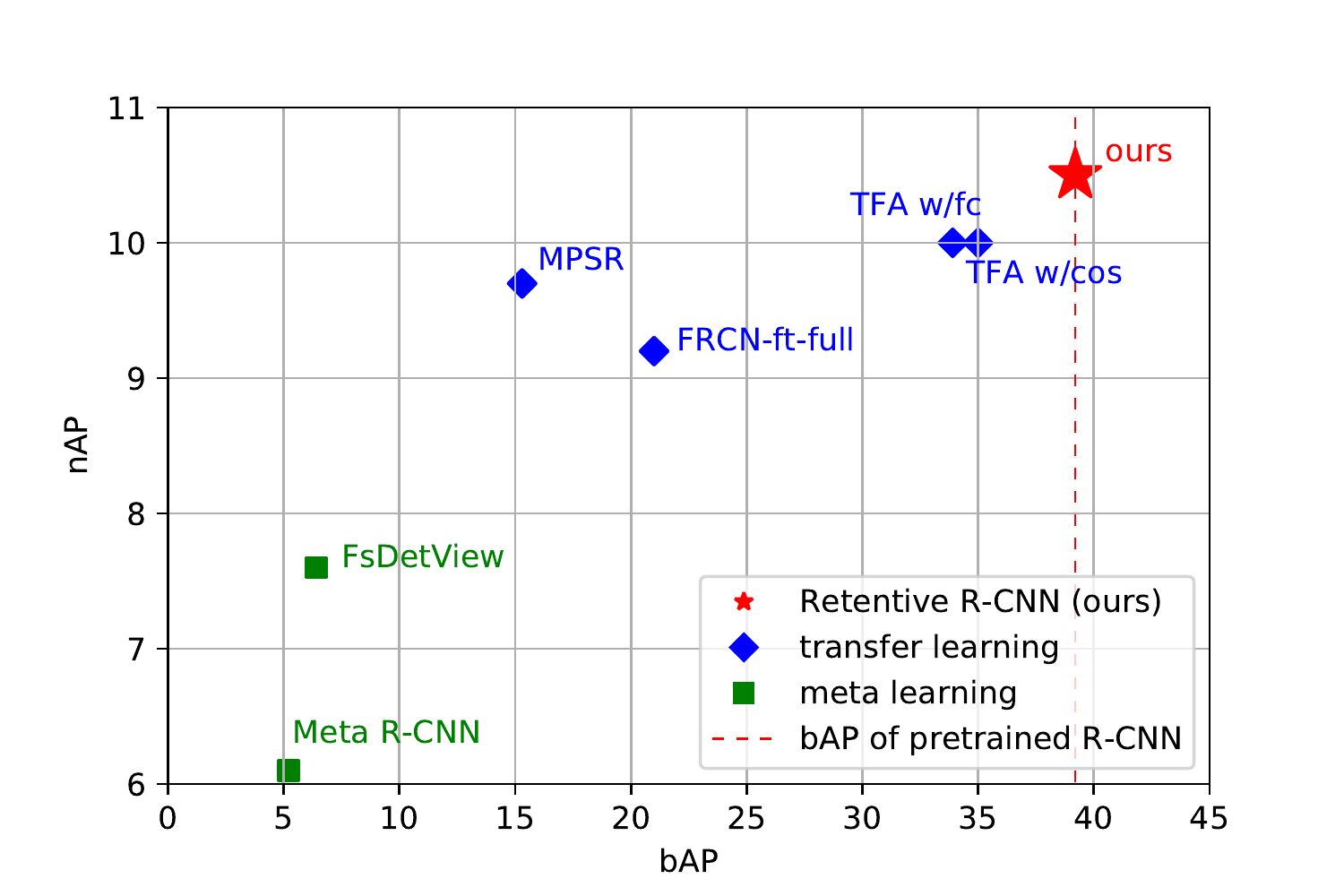}
\end{center}
   \caption{Performance of previous methods and ours on generalized few-shot objection on MS-COCO\cite{lin2014microsoft} under 10-shot settings, where base class AP and novel class AP are represented by x- and y-axis respectively. The red dashed line represents the base class AP of a base class detector. Our method does not degrade the base class AP while reaching state-of-the-art performance on novel categories.}
\vspace{-1em}
\label{fig:performance}
\end{figure}

Several recent works\cite{chen2018lstd,kang2019few-shot,yan2019meta,karlinsky2019repmet,wang2020frustratingly,wu2020multi-scale,xiao2020few-shot,hsieh2019one-shot,zhibo2020fgn,qi2019few-shot,juan-manuel2020incremental,wang2019meta-learning,michaelis2018one} have attempted to apply few-shot learning techniques on instance-level tasks, such as object detection, where an extra localization task is included and more complicated visual contexts and features encountered, making few-shot object detection way more challenging.
However, the majority focus merely on the performance of few-shot categories and ignore the catastrophic forgetting of base classes, which is not realistic.
Unlike image classification, the capability to detect the joint domain of both classes at once is even crucial for object detection since samples at test time may contain instances of both classes, which requires the detector to be computationally efficient and learn new concepts without catastrophic forgetting. 
The problem of detecting objects of both classes is called Generalized Few-Shot Detection (G-FSD).

A popular stream of few-shot object detection\cite{kang2019few-shot,yan2019meta,xiao2020few-shot,hsieh2019one-shot,qi2019few-shot} falls under the umbrella of meta-learning by leveraging external exemplars to do a visual search within the image. 
As their computational complexity is proportional to the number of categories, these methods become rather slow or even unavailable when tackling both sets of classes of a dataset.
A promising alternative is transfer learning based approaches\cite{chen2018lstd,yang2020context-transformer,wang2020frustratingly,wu2020multi-scale}, which can be trained incrementally to detect all classes in a single run. 
Wang \textit{et al.}\cite{wang2020frustratingly} share a similar interest in maintaining the overall performance on both classes and achieve competitive results by their two-stage finetuning approach (TFA), in which only the last layer of classification and box regression branch of RCNN\cite{girshick2014rich,girshick2015fast,ren2017faster} is finetuned while freezing backbone and RPN\cite{ren2017faster}.  
Nevertheless, there still exists a non-negligible base class performance gap with the pretrained model.

To diminish the gap, we first analyze the pretrained RCNN of TFA\cite{wang2020frustratingly} and find advantageous but neglected properties: 1) pretrained base class detector does not predict many false positives on novel class instances despite their saliency 2) RPN is biased on its seen classes instead of being ideally class-agnostic, thus freezing it without exposure to new classes can be suboptimal. 
By utilizing these properties, we propose a simple yet effective transfer learning based method, Retentive R-CNN, to meet the demands of G-FSD to learn without forgetting and detect all categories efficiently.
The name of Retentive R-CNN comes from its surprising ability to fully reserve the performance on base classes. Retentive R-CNN combines base and novel class detectors by Bias-Balanced RPN and Re-detector, introducing little extra cost.
Bias-Balanced RPN can better adapt to novel class objects and remain powerful on the base class, thus provides better proposals for both training and inference. 
Re-detector utilizes a consistency loss to regularize the adaptation during finetuning and takes advantage of the base class detector's property to incrementally detect without forgetting. 
It is worth mentioning that our method does not degrade the base class performance at all while achieving competitive performance on novel classes as well, as shown in Figure\ref{fig:performance}. Our contributions can be concluded as follows:

\begin{itemize}
    \item We find properties of base class detectors neglected in few-shot detection literature, which can be utilized to improve both base and novel class performance for transfer learning based methods with little overhead.
    \item We propose a few-shot detector without forgetting, Retentive R-CNN, with Bias-Balanced RPN and Re-detector to assist novel class adaptation with base class knowledge and ensemble base and novel class detectors.
    \item Our method achieves state-of-the-art overall performance on the few-shot detection benchmark\cite{wang2020frustratingly,kang2019few-shot} across all settings, with leading base class metrics and competitive novel class metrics.
\end{itemize}


\section{Related Work}
\textbf{Few-Shot Learning.} 
Previous few-shot learning literature mainly focuses on the task of image classification. 
Two popular approaches, metric learning\cite{vinyals2016matching, snell2017prototypical, koch2015siamese, sung2018learning} and meta-learning\cite{finn2017model, rusu2019meta}, have been widely adapted to avoid overfitting on the small data. 
Recent works\cite{chen2020new, wang2019simpleshot, dhillon2019baseline} also demonstrate the effectiveness of a pretrained backbone as a strong feature extractor and outperform many previous methods. 
However, catastrophic forgetting\cite{mccloskey1989catastrophic, lopez2017gradient} on base classes may happen during finetuning.
Gidaris \textit{et al.}\cite{gidaris2018dynamic} stress that a good few-shot learning system should adapt to new tasks rapidly while maintaining the performance on previous knowledge without forgetting\cite{mccloskey1989catastrophic, lopez2017gradient}, namely generalized few-shot learning, which is also the research interest of several other works\cite{qi2018low, qiao2018few,schonfeld2019generalized,li2019few}. It is worth mentioning that such an ability is more critical for object detection since images may have instances of both sets of categories.

\textbf{Object Detection.} 
Modern object detection has seen tremendous progress by utilizing deep convolutional networks. 
One of the representative architectures is R-CNN\cite{girshick2014rich,girshick2015fast,ren2017faster,he2017mask,lin2017feature}, which generates object proposals upon the holistic image features, then classify and refine the proposals given the features within it. R-CNN is also the architecture mostly explored in the context of few-shot object detection and the one we extend for G-FSD in this paper.
Impressive progress has also been made by single-stage methods\cite{redmon2016you,liu2016ssd,lin2017focal} and recent anchor-free methods\cite{law2019cornernet,duan2019centernet,tian2019fcos,shifeng2019bridging,borderdet,ma2021iqdet}.

\textbf{Few-Shot Object Detection.}
Exploration of few-shot object detection so far can be categorized into two streams: meta-learning\cite{kang2019few-shot,yan2019meta,hsieh2019one-shot,michaelis2018one,zhibo2020fgn,qi2019few-shot,karlinsky2019repmet,xiao2020few-shot,wang2019meta-learning,juan-manuel2020incremental} based and transfer learning based\cite{chen2018lstd,wang2020frustratingly,wu2020multi-scale,yang2020context-transformer}.
The majority of meta-learning stream predict detections conditioned on a set of support examples, which can be viewed as an exemplar-based visual search. For instance, Meta R-CNN\cite{yan2019meta} predicts upon ROI features reweighted by attentive vectors of each class, whose computational complexity grows linearly as the number of categories increases, making it hard to apply on large-scale datasets.
On the contrary, transfer learning based methods can easily employ full class detection. 
Transfer learning methods thus far have explored various aspects: Chen \textit{et al.}\cite{chen2018lstd} apply regularizations during finetuning, Yang \textit{et al.}\cite{yang2020context-transformer} utilize a non-local structure to model global context, Wu \textit{et al.}\cite{wu2020multi-scale} augment training samples to mitigate scale bias due to limited data.

A handful of works share a similar focus on well detecting both classes: Juan-Manuel \textit{et al.}\cite{juan-manuel2020incremental} try to tackle it with a meta-learned CenterNet\cite{duan2019centernet}, though the performance is still limited with the linearly growing complexity issue;
Wang \textit{et al.}\cite{wang2020frustratingly} propose TFA with a simple pretrain-finetune scheme for G-FSD.
In TFA, adapting to novel classes with less degradation on base classes is achieved by two-stage finetuning: first finetune on novel classes, then use the weights of novel classes as initialization to finetune on both classes. 
The metrics on base classes are somewhat reserved by this procedure and a slow learning schedule.
Nevertheless, the performance drop in base classes still exists.

\section{Approach}
In this section, we start with the problem formulation of few-shot object detection. 
Next, we investigate the representative transfer learning based TFA\cite{wang2020frustratingly} to reveal some neglected properties of the pretrained base detector. 
Then we describe our proposed model, which utilizes these properties, followed by training and inference details.

\subsection{Problem Statement}
Following previous literature\cite{kang2019few-shot,wang2020frustratingly}, we split the categories of a dataset into base classes $\mathcal{C}_b$ and novel classes $\mathcal{C}_n$, with $\mathcal{D}_b$ and $\mathcal{D}_n$ denoting the corresponding sub-datasets, respectively. 
$\mathcal{D}_b$ contains abundant annotations for training, while only a few data in $\mathcal{D}_n$ is available. 
Our objective is to learn a detection model $f(\cdot)$ for both $\mathcal{C}_b$ and $\mathcal{C}_n$ from the few novel class samples without forgetting the learned capability from the abundant base class samples.

Such an objective can be easily achieved by meta training a model to perform an exemplar-based visual search on $\mathcal{D}_b$, then directly deploy it without finetuning, as in one-shot detection literature \cite{hsieh2019one-shot,michaelis2018one}. 
However, these methods do not perform as good as ordinary detection methods on $\mathcal{D}_b$ and require high time and space complexity. 
On the contrary, transfer learning based methods can efficiently deal with full-way detection and achieve competitive results on $\mathcal{C}_n$, as demonstrated in previous works\cite{wang2020frustratingly,wu2020multi-scale}. 
Thus we propose to tackle the problem of G-FSD in a transfer learning paradigm: first obtain a base model $f^b$ by training on $\mathcal{D}_b$ and then obtain a novel model $f^n$ via finetuning $f^b$ on $\mathcal{D}_n$ (or a combination of $\mathcal{D}_b$'s subset and $\mathcal{D}_n$). 
However, the finetuning stage tends to degrade the base class performance due to the forgetting effect\cite{mccloskey1989catastrophic,lopez2017gradient} if it is finetuned on $\mathcal{D}_n$, or due to the sample limitation on $\mathcal{D}_b$ to balance class frequency if finetuned on both classes. Regarding this problem, a question is probably raised: is the degradation unavoidable?

\begin{figure}[!ht]
	\centering
	\includegraphics[width = 7.5cm]{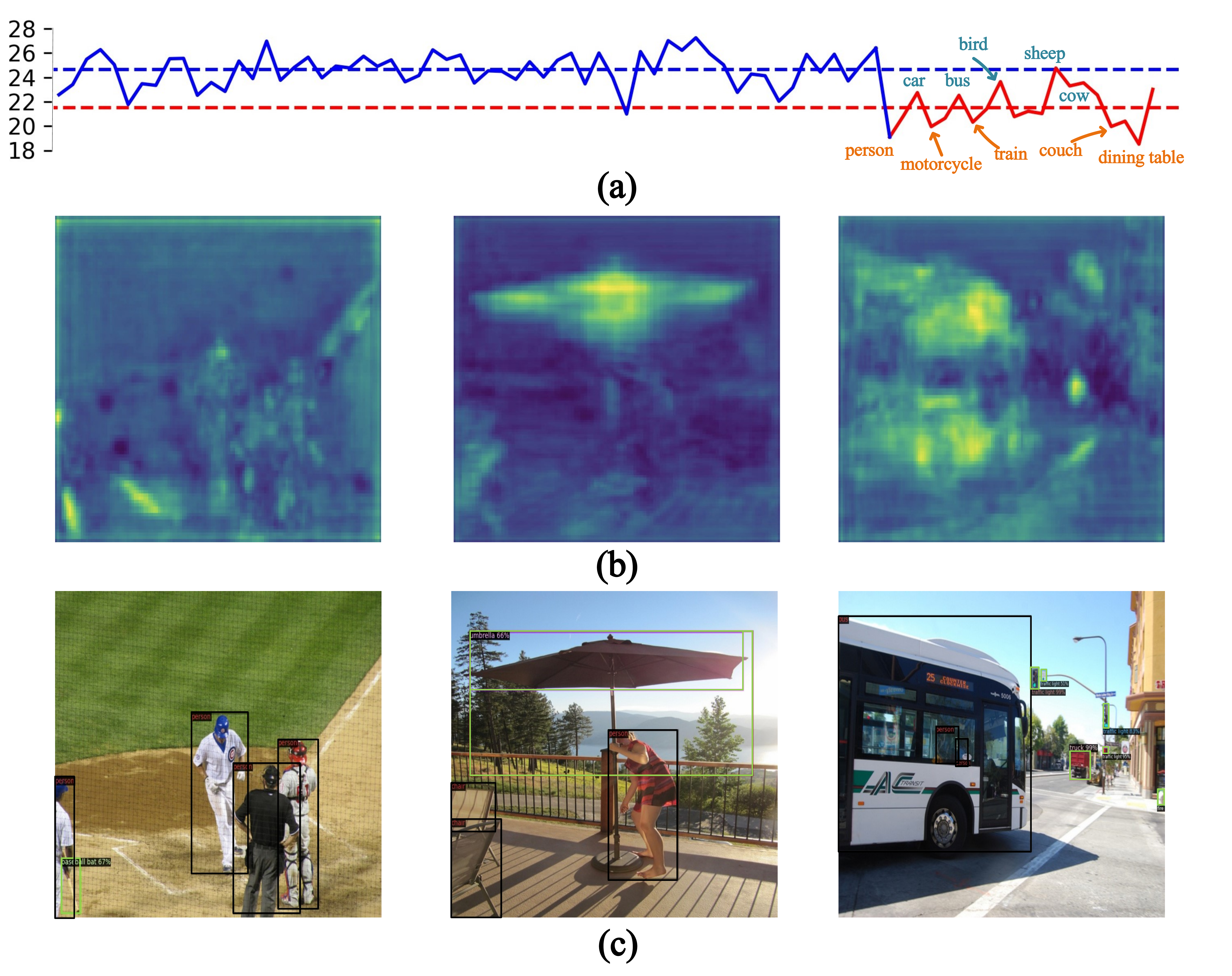}
	\caption{(a) L2-norms of ROI features extracted from a pretrained R-CNN on $\mathcal{D}^b$, sorted by class index. Blue and red represent seen and unseen classes, respectively, and dashed lines denote their average. Unseen classes with norms significantly higher or lower than average are annotated with class names in blue or red, respectively . (b) Backbone features' (FPN-P3) L2-norm map and (c) the corresponding detections of the base detector (green boxes) and ground truths of unseen classes (black boxes). The base detector has a strong ability to reject unseen classes.}
	\vspace{-20pt}
\label{fig:norm-and-det}
\end{figure}

\begin{table}[!ht]
\small
    \centering
    \begin{tabular}{ c c c c}
        \Xhline{1pt}
          \textbf{Metric / Component} & RPN & RCNN & drop\\\hline
        \textbf{uAR@1000} & 34.1 & 8.5 & \textbf{25.6}\\
        \textbf{AR@1000} & 61.1 & 54.7 & 6.4 \\\Xhline{1pt}
    \end{tabular}
    \caption{Recall between the output of RPN, the detector and the ground truths of unseen classes (uAR) and seen classes (AR).}
    \label{tab:unseen-ar}
\end{table}

\begin{table}[!ht]
\small
    \centering
    \begin{tabular}{ c c c}
        \Xhline{1pt}
          \textbf{Metric / RPN} & pretrained & finetuned\\\hline
        \textbf{AR@100} & 31.3 & \textbf{34.2}\\
        \textbf{AR@1000} & 45.8 & \textbf{48.0}\\\Xhline{1pt}
    \end{tabular}
    \caption{Mean average recall between the output of RPN and ground truths of both base and novel classes.}
    \vspace{-1em}
    \label{tab:rpn-ar}
\end{table}

\subsection{Analysis on Transfer Learning based Few-Shot Object Detection}
\label{sec:analysis}
To answer this question without loss of generality, we analyze TFA\cite{wang2020frustratingly}'s properties as a representative transfer learning model on few-shot detection tasks.
TFA is first pretrained on $\mathcal{D}_b$ as ordinary R-CNN, then the last layers in classification and box regression heads are tuned on $\mathcal{D}_n$. 
The finetuned novel class heads' weights are concatenated with base class weights as the initialization for the final finetuning on a combined dataset consists of $\mathcal{D}_n$ and $\mathcal{D}_b$'s subset, where the number of samples per category is enforced to be identical. A slow and steady learning schedule is also applied during the final finetuning stage.
Take 10-shot settings on MS-COCO for example, AP on $\mathcal{C}_b$ is better reserved than a pure finetuning baseline (31.8 to 35.0), though AP of the base class detector can achieve as high as 39.2.

\begin{figure*}[!ht]
	\centering
	\includegraphics[width = 17.5cm]{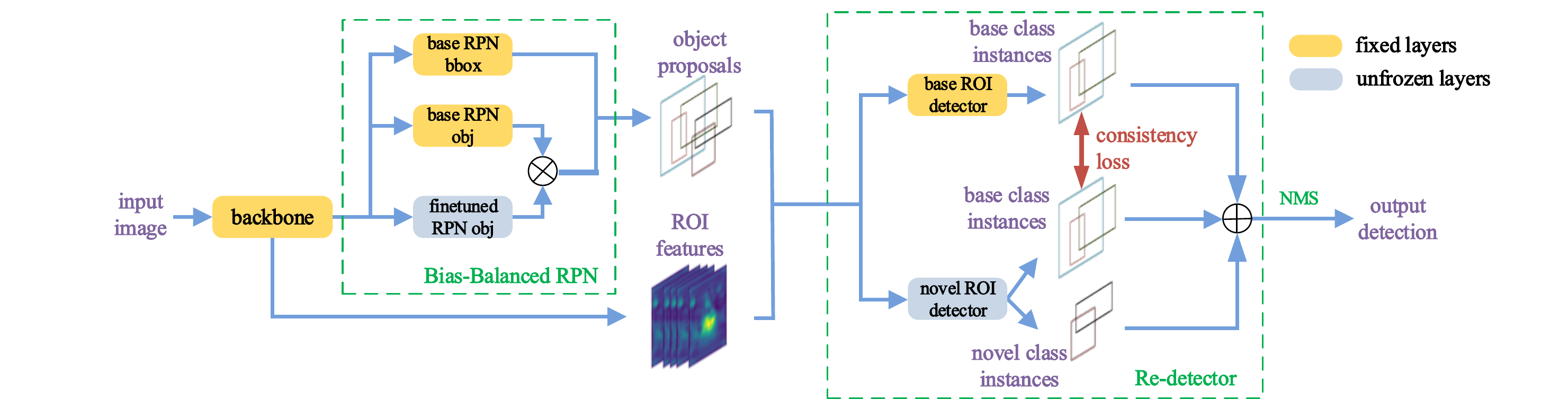}
	\caption{An overview of the proposed Retentive R-CNN. We implement Bias-Balanced RPN to debias the pretrained RPN and Re-detector to detect objects of both classes without forgetting, where a consistency loss is utilized to regularize finetuning. $\otimes$ represents ensembling operation, which is max in our implementation.}
	\vspace{-10pt}
\label{fig:overview}
\end{figure*}

\textbf{Why cosine classifier works?} 
Cosine classifier is commonly adapted in few-shot classification\cite{gidaris2018dynamic,qi2018low} as cosine similarity bridged transfer learning and metric learning approach and generally performs well on base and novel class trade-off.
The conclusion remains valid for TFA\cite{wang2020frustratingly} as the base class performance is generally higher with a cosine classifier. We collect the ROI features from an R-CNN pretrained on $\mathcal{D}_b$ of MS-COCO\cite{lin2014microsoft} and compute the average pixel-wise L2-norm of $\mathcal{C}_b$ and $\mathcal{C}_n$. 
The results are shown in Figure\ref{fig:norm-and-det}(a). A massive variation of norms between base classes and unseen novel classes can be easily observed. This may account for the effectiveness of cosine classifiers for being agnostic to feature norms. Also, the norms of unseen classes with closer relationship with seen classes are relatively higher (blue names annotated in Figure\ref{fig:norm-and-det}(a)).

\textbf{Does base detector find novel class salient objects?}
To a large extent, no. 
We hypothesize it is due to the features deactivated during training on $\mathcal{C}_b$ as indicated by low L2-norm, which will not produce a high confidence score with a dot-product classifier. 
We visualize the detection results and its feature norms on FPN\cite{lin2017feature} P3 in Figure\ref{fig:norm-and-det}. 
The local features around the people in the first two images are obviously deactivated, although the object is of great saliency to humans. 
Features of the bus are somewhat activated in the third image of Figure\ref{fig:norm-and-det}(b) (probably due to close relationship with trucks of $\mathcal{C}_b$), yet the detector is still able to recognize it as background. 
More results indicating this property are provided in the supplementary materials without cherry-picking.
To quantitatively answer this question, the average recall between $f^b$'s RPN proposals, final outputs, and ground truths of unseen class $\mathcal{C}_n$ (uAR) and seen classes (AR), is calculated in Table\ref{tab:unseen-ar}. 
The drastic drop of uAR well demonstrates the ability of $f^b$ to reject novel class objects. 
Thus we can utilize this property to reserve base class performance as $f^b$ does not introduce many false positives on $\mathcal{C}_b$ when encountering novel class instances.

\textbf{Is RPN class-agnostic?}
While most transfer learning\cite{wang2020frustratingly,wu2020multi-scale} and meta-learning\cite{yan2019meta,xiao2020few-shot,karlinsky2019repmet} works treat RPN as class-agnostic and freeze it during finetuning, RPN is not ideally class-agnostic and biased on its seen categories.
During training on $\mathcal{D}_b$, anchors of novel class instances are categorized into non-object due to lack of annotations, making RPN bias on training samples.
We compare AR of \textit{all} classes of a finetuned RPN on $\mathcal{C}_b\cup \mathcal{C}_n$ under 10-shot setting with pretrained RPN in Table\ref{tab:rpn-ar}, where the apparent improvement validates our answer.

\subsection{Retentive R-CNN}
\label{sec:retentive}
Our proposed model for G-FSD, Retentive R-CNN, consists of Bias-Balanced RPN and Re-detector to utilize the aforementioned properties of the base class detector $f^b$. The model architecture is illustrated in Figure\ref{fig:overview}. 

\textbf{Re-detector.}
Re-detector consists of two detector heads, predicting detections of $\mathcal{C}_b$ and $\mathcal{C}_b\cup \mathcal{C}_n$ from object proposals in parallel, where one stream remains the same weight as in $f^b$ to predict objects of $\mathcal{C}_b$ (denote as $det^b$) and the other holds the finetuned weights to detect objects of both $\mathcal{C}_n$ and $\mathcal{C}_{b}$ (denote as $det^n$). 
Detecting both classes can well alleviate the false positives due to inadequate data training, as shown in Section\ref{sec:ablation}.
$det^b$ utilize a fully-connected layer for classification and $det^n$ use a cosine classifier to balance the variation of features in their norms.
Similar to TFA, we finetune merely the last layers of classification and box regression head of $det^n$, which is capable of producing competitive results.

As $f^b$ is trained from abundant data, we hope that $det^n$ can inherit the reliable knowledge of $f^b$. Towards this end, we propose an auxiliary consistency loss to regularize $det^n$ to score object proposals similar to $det^b$ on the base class entries, which takes the form of KL-Divergence as in previous knowledge distillation works\cite{hu2016deep,yu2017visual}. For proposals of $C_b$, $det^n$ is enforced to predict high confidence, and for proposals not belonging to $\mathcal{C}_b$, $det^n$ mimics $det^b$ with similarly low probabilities. Given the final probabilities $p_c^b$ and $p_c^n$ of class $c$ predicted by $det^b$ and $det^n$, the consistency loss is formalized as:

\begin{equation}
    \mathcal{L}_{con}=
    \sum_{c\in\mathcal{C}_b}
    \Tilde{p_c^n} log(\frac{\Tilde{p_c^n}}{\Tilde{p_c^b}})
\end{equation}
where $\Tilde{p_i^n}=\frac{p_i^n}{\sum_{c\in\mathcal{C}_b} p_c^n}$ and the same for $\Tilde{p_i^b}$. 
This is quite different from TK in LSTD\cite{chen2018lstd} where the KL-Divergence is computed between the highest probabilities of $\mathcal{C}_b$ and $\mathcal{C}_n$.
Note that $p_i^n$ is the normalized marginal probability distribution over base classes after softmax over all class entries. The total loss of Re-detector during finetuning stage is

\begin{equation}
    \mathcal{L}_{det}=\mathcal{L}^n_{cls}+\mathcal{L}^n_{box}+\lambda \mathcal{L}_{con}
\end{equation}
where $\mathcal{L}_{cls}$ and $\mathcal{L}_{box}$ takes the same form as Faster R-CNN\cite{ren2017faster} and is computed on $det^n$ only, and $\lambda$ denotes the coefficient for the consistency loss.

\textbf{Bias-Balanced RPN.} 
R-CNN relies on RPN to generate object proposals as training samples for second stage classification and other subsequent processing.
The quality of RPN proposals is especially crucial when the network is trained under low-data scenarios.
As shown in Section\ref{sec:analysis}, a pretrained RPN may fail to catch novel class objects, further aggravating the scarcity of samples, while a finetuned RPN can alleviate this issue, thus providing better samples for second-stage modules to learn. 
We try to unfreeze different layers of RPN for finetuning and empirically, unfreeze the final layer that predicts objectness is sufficient to produce a noticeable improvement (results given in Setion\ref{sec:ablation}). 

To retain performance on base classes, we propose Bias-Balanced RPN to integrate both pretrained RPN and the finetuned one. 
It ensembles the objectness prediction heads to raise $\mathcal{C}_b$ and $\mathcal{C}_n$ proposals properly. 
Given a feature map of size $H\times W$, base RPN predicts an objectness map $\mathcal{O}^{H\times W}_b$ and finetuned RPN predicts $\mathcal{O}^{H\times W}_n$, the final output objectness of Bias-Balanced RPN is defined as $\mathcal{O}^{H\times W}=\max(\mathcal{O}^{H\times W}_b,\mathcal{O}^{H\times W}_n)$. 
Note that, during the finetuning stage, only the objectness of finetuned RPN is set unfrozen. 
Box regression and the convolution layer are shared across base RPN and finetuned RPN, as illustrated in Figure\ref{fig:overview}. 
Theoretically, the max operation guarantees the RPN not to overlook proposals of previously learned classes catastrophically.
With little computational overhead and extra weights, we believe Bias-Balanced RPN can serve as a general component for G-FSD.
The full loss function of Retentive R-CNN during the finetuning stage is

\begin{equation}
    \mathcal{L}_{ft}=\mathcal{L}^n_{obj}+\mathcal{L}_{det}
\end{equation}
where $\mathcal{L}^n_{obj}$ is the binary cross-entropy loss on finetuned RPN's objectness layer.

\begin{table*}
    \centering
    \begin{tabular}{c c | c c c | c c c | c c c}
      \Xhline{1pt}
      \multicolumn{2}{c |}{\multirow{2}{*}{\textbf{Methods} / \textbf{Shots}}} &
      \multicolumn{3}{c |}{\textbf{5 shot}}  &
      \multicolumn{3}{c |}{\textbf{10 shot}}  &
      \multicolumn{3}{c}{\textbf{30 shot}}\\
    \multicolumn{2}{c |}{} & \textbf{AP} &  \textbf{bAP} & \textbf{nAP} & \textbf{AP} &  \textbf{bAP} & \textbf{nAP} & \textbf{AP} &  \textbf{bAP} & \textbf{nAP} \\\hline
      Ours & Retentive R-CNN & \best{31.5} & \best{39.2} & \second{8.3} & \best{32.1} & \best{39.2} & \best{10.5} & \best{32.9} & \best{39.3} & \second{13.8} \\\hline
      \multirow{5}{*}{Transfer Learning} 
      & FRCN-ft-full\cite{wang2020frustratingly} & 18.0 & 22.0 & 6.0 & 18.1 & 21.0 & 9.2 & 18.6 & 20.6 & 12.5 \\
      & FRCN-BCE & \second{29.1} & \second{36.8} & 6.0 & \second{29.2} & \second{36.8} & 6.4 & 30.2 & \second{36.8} & 10.3 \\
      & TFA w/ fc\cite{wang2020frustratingly} & 27.5 & 33.9 & \best{8.4} & 27.9 & 33.9 & \second{10.0} & 29.7 & 35.1 & 13.4 \\
      & TFA w/ cos\cite{wang2020frustratingly} & 28.1 & 34.7 & 8.3 & 28.7 & 35.0 & \second{10.0} & \second{30.3} & 35.8 & 13.7 \\
      & MPSR\cite{wu2020multi-scale} & - & - & - & 15.3 & 17.1 & 9.7 & 17.1 & 18.1 & \best{14.1} \\\hline
      \multirow{4}{*}{Meta Learning} & ONCE~\cite{juan-manuel2020incremental} & 13.7 & 17.9 & 1.0 & 13.7 & 17.9 & 1.2 & - & - & -  \\
       & Meta R-CNN$^{*}$\cite{yan2019meta} & 3.6 & 3.5 & 3.8 & 5.4 & 5.2 & 6.1 & 7.8 & 7.1 & 9.9 \\
       & FSRW\cite{kang2019few-shot} & - & - & - & - & - & 5.6 & - & - & 9.1 \\
       & FsDetView$^{*}$\cite{xiao2020few-shot} & 5.9 & 5.7 & 6.6 & 6.7 & 6.4 & 7.6 & 10.0 & 9.3 & 12.0 \\\Xhline{1pt}
    \end{tabular}
    \caption{Few-shot object detection results on MS-COCO under 5,10,30-shot settings, best viewed in color. AP, bAP, nAP represents mAP of MS-COCO for all classes, base classes, and novel classes, respectively. Best results and second-best are colored in {\color{red}red} and {\color{blue}blue}, respectively, ‘-’ means the result is not reported in the original paper. We outperform or on-par with all previous methods for each metric under these settings, with significant improvements on AP and bAP.}
    \label{tab:coco}
\end{table*}

\begin{table*}
\footnotesize
    \centering
    \begin{tabular}{p{0.9cm} c | c c c c c | c c c c c | c c c c c}
      \Xhline{1pt}
      \multicolumn{2}{c |}{\multirow{2}{*}{\textbf{Methods} / \textbf{Shots}}} &
      \multicolumn{5}{c |}{\textbf{All Set 1}}  &
      \multicolumn{5}{c |}{\textbf{All Set 2}}  &
      \multicolumn{5}{c}{\textbf{All Set 3}}\\
    \multicolumn{2}{c |}{} & 1 &  {2} & {3} & {5} &  {10} & 1 &  {2} & {3} & {5} &  {10} & 1 &  {2} & {3} & {5} &  {10} \\\hline
      \ \ Ours & Retentive R-CNN & {\color{red}\textbf{71.3}} & {\color{red}\textbf{72.3}} & {\color{red}\textbf{72.1}} & {\color{red}\textbf{74.0}} & {\color{red}\textbf{74.6}} & {\color{red}\textbf{66.8}} & {\color{red}\textbf{68.4}} & {\color{red}\textbf{70.2}} & {\color{red}\textbf{70.7}} & {\color{red}\textbf{71.5}} & {\color{red}\textbf{69.0}} & {\color{red}\textbf{70.9}} & {\color{red}\textbf{72.3}} & {\color{red}\textbf{73.9}} & {\color{red}\textbf{74.1}} \\\hline
      \multirow{4}{*}{\thead{Transfer\\Learning}} 
      & FRCN-ft-full\cite{wang2020frustratingly} & 55.4 & 57.1 & 56.8 & 60.1 & 60.9 & 50.1 & 53.7 & 53.6 & 55.9 & 55.5 & 58.5 & 59.1 & 58.7 & 61.8 & 60.8 \\
      & TFA w/ fc\cite{wang2020frustratingly} & 69.3 & 66.9 & 70.3 & \second{73.4} & \second{73.2} & 64.7 & \second{66.3} & \second{67.7} & \second{68.3} & \second{68.7} & 67.8 & \second{68.9} & 70.8 & 72.3 & 72.2 \\
      & TFA w/ cos\cite{wang2020frustratingly} & \second{69.7} & \second{68.2} & \second{70.5} & \second{73.4} & 72.8 & \second{65.5} & 65.0 & \second{67.7} & 68.0 & 68.6 & \second{67.9} & 68.6 & \second{71.0} & \second{72.5} & \second{72.4} \\
      & MPSR\cite{wu2020multi-scale} & 56.8 & 60.4 & 62.8 & 66.1 & 69.0 &
53.1 & 57.6 & 62.8 & 64.2 & 66.3 & 55.2 & 59.8 & 62.7 & 66.9 & 67.7 \\\hline
      \multirow{3}{*}{\thead{Meta\\Learning}} &
        Meta R-CNN$^{*}$\cite{yan2019meta} & 17.5 & 30.5 & 36.2 & 49.3 & 55.6 & 19.4 & 33.2 & 34.8 & 44.4 & 53.9 & 20.3 & 31.0 & 41.2 & 48.0 & 55.1 \\
       & FSRW\cite{kang2019few-shot} & 53.5 & 50.2 & 55.3 & 56.0 & 59.5 & 55.1 & 54.2 & 55.2 & 57.5 & 58.9 & 54.2 & 53.5 & 54.7 & 58.6 & 57.6 \\
       & FsDetView$^*$\cite{xiao2020few-shot} & 36.4 & 40.3 & 40.1 & 50.0 & 55.3 & 36.3 & 43.7 & 41.6 & 45.8 & 54.1 & 37.0 & 39.5 & 40.7 & 50.7 & 54.8 \\\Xhline{1pt}
    \end{tabular}
    \caption{Few-shot object detection results on Pascal VOC(07+12) \textbf{all classes (AP$\mathbf{_{50}}$)} under 1,2,3,5,10-shot settings, best viewed in color. Best results and second-best are colored in {\color{red}red} and {\color{blue}blue}, respectively. Thanks to the non-forgetting ability, Retentive R-CNN consistently outperforms other methods w.r.t overall AP under all the data settings.}
    \label{tab:voc-all}
\end{table*}

\textbf{Training.}
As a transfer learning based method, Retentive R-CNN is trained in two stages: pretraining on $\mathcal{D}_b$ and then finetune on the combined dataset of $\mathcal{D}_n$ and $\mathcal{D}_b$'s subset. 
As aforementioned, we only unfreeze three layers: objectness of the finetuned RPN, the last linear layers of classification and box regression of $det^n$.
Thanks to the capability of retaining base class performance, we can apply a swifter learning schedule for finetuning, \textit{e.g.}, 5000 iterations for 10-shot MS-COCO\cite{lin2014microsoft} compared to 160000 iterations of TFA\cite{wang2020frustratingly}.

\textbf{Inference.}
Given the object proposals from Bias-Balanced RPN, the corresponding features are fed into the two heads of Re-detector in parallel. 
The set of predicted boxes of both heads are gathered into one for the final NMS procedure.
As $det^b$ is somehow more reliable as it learns from abundant data, we add a little bonus (0.1 in our implementation) for the scores predicted from $det^b$ if they surpass the pre-NMS threshold, which could encourage the NMS procedure to take $det^b$'s output when $det^b$ and $det^n$ find similar base class results.
More details will be described in the supplementary material.
As the backbone and feature transformation layers in Bias-Balanced RPN and Re-detector are shared among both detector heads, we can maintain the base class performance with little overhead compared to an ordinary R-CNN.


\section{Experiments}
\subsection{Experimental Settings}
We evaluate our method on the well-established few-shot detection benchmark\cite{wang2020frustratingly,kang2019few-shot} based on MS-COCO\cite{lin2014microsoft} and Pascal VOC \cite{everingham2010pascal}, following \textbf{the same class splits and data splits} in previous works\cite{kang2019few-shot,wang2020frustratingly,wu2020multi-scale} for a fair comparison.  
We report 5,10,30-shot results on MS-COCO and 1,2,3,5,10-shot results on 3 random splits of Pascal VOC. 
Towards the problem of G-FSD, \textbf{the overall performance of both classes is our major concern}.
We reproduce Meta R-CNN\cite{yan2019meta} and FsDetView\cite{xiao2020few-shot} using exactly the same samples for finetuning without hyperparameter changing (by running their official code) and denote the reproduced results with a * at the upper right corner.
Results for ONCE\cite{juan-manuel2020incremental}, MetaDet\cite{wang2019meta-learning} and FSRW\cite{kang2019few-shot} are reported from their original paper.

We use an ImageNet pretrained ResNet-101\cite{he2016deep} with FPN\cite{lin2017feature} as the backbone. 
Pretraining on $\mathcal{D}_b$ is the same as in \cite{wang2020frustratingly}, then the finetuning layers are initialized by random. For all experiments, we set learning rate to 0.05 and $\lambda$ to 0.1 to finetune until full convergence.

\subsection{Comparison Experiments}
We compared our results with both transfer learning\cite{wang2020frustratingly,wu2020multi-scale} and meta-learning based methods\cite{juan-manuel2020incremental,yan2019meta,kang2019few-shot,xiao2020few-shot}. 
Towards maintaining base class performance, one can quickly come up with an R-CNN model with N binary classifiers for detecting a dataset of N classes as binary classifiers are decoupled with each other.
We also train such a model (denoted as FRCN-BCE) with binary cross-entropy loss for ROI classification as a strong baseline, using the same hyperparameters as Retentive R-CNN except for initializing the classifiers' bias as in RetinaNet\cite{lin2017focal}. 

\textbf{Results on MS-COCO\cite{lin2014microsoft}.}
Table\ref{tab:coco} shows mean average precision over 0.5 to 0.95 IOU thresholds on all, base, and novel classes (AP, bAP, nAP) under different data settings. 
We outperform previous methods significantly on AP and bAP, as our method does not degrade on base classes at all. 
Meanwhile, we achieve competitive results on novel classes as well (state-of-the-art for 10-shot and on-par with state-of-the-art for 5- and 30-shot). 

Towards the same objective to incrementally detect rare objects, ONCE\cite{juan-manuel2020incremental} does not degrade the base class performance as well, yet its performance on both classes is limited.
The very competitive TFA\cite{wang2020frustratingly} models can gradually recover base class performance with samples increasing; however, the gap is still indispensable, \textit{e.g.}, the bAP gap between 30-shot TFA w/cos\cite{wang2020frustratingly} and the base model is as large as 3.4.
As expected, FRCN-BCE can maintain base class performance from its pretrained model intrinsically, but the performance on both base and novel class is lower than an ordinary RCNN by a large margin.
Given that Retentive R-CNN only adds little overhead with layers mostly shared, our method is a superior choice for G-FSD.
Despite MPSR\cite{wu2020multi-scale} slightly outperform our method with respect to nAP on 30-shot, the performance drop on base classes is significant, and thus it is not suitable for G-FSD. 
An even larger performance drop can be observed in Meta R-CNN\cite{yan2019meta} and FsDetView\cite{xiao2020few-shot}, probably because they predict the whole probability distribution of an ROI from features reweighted by a certain class-attentive vector. 
The vast performance drop is alleviated to some extent in FSRW\cite{kang2019few-shot} (see Table\ref{tab:voc-all}), which only predicts the probability for the class of the reweighting vector. 

\textbf{Results on Pascal-VOC\cite{everingham2010pascal}.}
Table\ref{tab:voc-all} and Table\ref{tab:voc-novel} show overall and novel class results on VOC benchmark respectively. 
Results of Meta R-CNN\cite{yan2019meta} from original paper are also included in Table\ref{tab:voc-novel} as a reference. 
Note that the results are not directly comparable because samples used for finetuning are different, which can make a significant impact on the final metrics.
We consistently outperform all methods on overall AP across all datasplits thanks to the non-forgetting property. 
As stated above, MPSR\cite{wu2020multi-scale} and several other meta-learning methods\cite{yan2019meta,xiao2020few-shot,kang2019few-shot} do not perform well on overall performance as the base class knowledge is forgotten during the finetuning stage. 

Notice that performance on novel classes is not our primary concern, though, competitive results are achieved by Retentive R-CNN under most cases on VOC novel classes as shown in Table\ref{tab:voc-novel}. 
MPSR\cite{wu2020multi-scale} made most of the best nAP records; however, non-negligible base class performance is sacrificed. Compared to the methods that better preserves base class performance, we outperform the current best TFA\cite{wang2020frustratingly} in most cases, with approximative results under the rest.

\begin{table*}
\footnotesize
    \centering
    \begin{tabular}{p{0.9cm} c | c c c c c | c c c c c | c c c c c}
      \Xhline{1pt}
      \multicolumn{2}{c |}{\multirow{2}{*}{\textbf{Methods} / \textbf{Shots}}} &
      \multicolumn{5}{c |}{\textbf{Novel Set 1}}  &
      \multicolumn{5}{c |}{\textbf{Novel Set 2}}  &
      \multicolumn{5}{c}{\textbf{Novel Set 3}}\\
    \multicolumn{2}{c |}{} & 1 &  {2} & {3} & {5} &  {10} & 1 &  {2} & {3} & {5} &  {10} & 1 &  {2} & {3} & {5} &  {10} \\\hline
      \ \ Ours & Retentive R-CNN & \second{42.4} & \best{45.8} & \second{45.9} & 53.7 & 56.1 & 21.7 & 27.8 & \second{35.2} & \second{37.0} & \second{40.3} & 30.2 & \second{37.6} & \second{43.0} & \best{49.7} & 50.1 \\\hline
      \multirow{4}{*}{\thead{Transfer\\Learning}} 
      & FRCN-ft-full\cite{wang2020frustratingly} & 15.2 & 20.3 & 29.0 & 25.5 & 28.7 & 13.4 & 20.6 & 28.6 & 32.4 & 38.8 & 19.6 & 20.8 & 28.7 & 42.2 & 42.1 \\
      & TFA w/ fc\cite{wang2020frustratingly} & 36.8 & 29.1 & 43.6 & \best{55.7} & \second{57.0} & 18.2 & \best{29.0} & 33.4 & 35.5 & 39.0 & 27.7 & 33.6 & 42.5 & 48.7 & \second{50.2} \\
      & TFA w/ cos\cite{wang2020frustratingly} & 39.8 & 36.1 & 44.7 & \best{55.7} & 56.0 & \second{23.5} & 26.9 & 34.1 & 35.1 & 39.1 & 30.8 & 34.8 & 42.8 & 49.5 & 49.8 \\
      & MPSR\cite{wu2020multi-scale} & \best{42.8} & \second{43.6} & \best{48.4} & \second{55.3} & \best{61.2} &
\best{29.8} & \second{28.1} & \best{41.6} & \best{43.2} & \best{47.0} & \best{35.9} & \best{40.0} & \best{43.7} & \second{48.9} & \best{51.3} \\\hline
      \multirow{5}{*}{\thead{Meta\\Learning}} &
      Meta R-CNN\cite{yan2019meta} & 19.9 & 25.5 & 35.0 & 45.7 & 51.5 & 10.4 & 19.4 & 29.6 & 34.8 & 45.4 & 14.3 & 18.2 & 27.5 & 41.2 & 48.1 \\
      & Meta R-CNN$^{*}$\cite{yan2019meta} & 16.8 & 20.1 & 20.3 & 38.2 & 43.7 & 7.7 & 12.0 & 14.9 & 21.9 & 31.1 & 9.2 & 13.9 & 26.2 & 29.2 & 36.2 \\
       & FSRW\cite{kang2019few-shot} & 14.8 & 15.5 & 26.7 & 33.9 & 47.2 & 15.7 & 15.3 & 22.7 & 30.1 & 39.2 & 19.2 & 21.7 & 25.7 & 40.6 & 41.3 \\
       & MetaDet\cite{wang2019meta-learning} & 18.9 & 20.6 & 30.2 & 36.8 & 49.6 & 21.8 & 23.1 & 27.8 & 31.7 & 43.0 & 20.6 & 23.9 & 29.4 & 43.9 & 44.1\\
       & FsDetView$^*$\cite{xiao2020few-shot} & 25.4 & 20.4 & 37.4 & 36.1 & 42.3 & 22.9 & 21.7 & 22.6 & 25.6 & 29.2 & \second{32.4} & 19.0 & 29.8 & 33.2 & 39.8 \\\Xhline{1pt}
    \end{tabular}
    \caption{Few-shot object detection results on Pascal VOC(07+12) \textbf{novel classes (nAP$\mathbf{_{50}}$)} under 1,2,3,5,10-shot settings, best viewed in color. Best results and second-best are colored in {\color{red}red} and {\color{blue}blue}, respectively. Although nAP is \textbf{not} our primary concern, our method makes competitive results. MPSR\cite{wu2020multi-scale} made most of the best nAP; however, base class metrics are largely sacrificed, as shown in Table\ref{tab:voc-all}. We outperform TFA\cite{wang2020frustratingly}, the current best method paying fair attention to both classes, in most cases and on-par with it in the rest.}
    \label{tab:voc-novel}
\end{table*}

\subsection{Ablation Study and Visualization}
\label{sec:ablation}
Without loss of generality, we conduct ablation experiments on COCO benchmark under 10-shot scenario. All models are trained with the same hyperparameters unless otherwise stated.

\begin{table}[t]
\small
    \centering
    \begin{tabular}{ c c |c c c c}
        \Xhline{1pt}
          \textbf{cls} & \textbf{bbox} & \textbf{AR} & \textbf{AP} & \textbf{bAP} & \textbf{nAP} \\\hline
        \textbf{max} & \textbf{-} & \textbf{47.8} & \textbf{32.1} & 39.2 & \textbf{10.5}\\
         - & - & 45.6 & 32.0 & \textbf{39.3} & 10.1 \\
        arith-avg & - & 47.1 & 32.0 & \textbf{39.3} & 10.3\\
        geo-avg & - & 33.5 & 30.5 & 37.4 & 9.6\\
        max & unfreeze & 47.7 & 32.0 & \textbf{39.3} & 10.4\\
        max & arith-avg & 47.7 & 32.0 & 39.2 & 10.4\\\Xhline{1pt}
    \end{tabular}
    \caption{AR and AP results for base and novel classes among different RPN variants. Results of RPN components are ensembled by functions, including max, arithmetic average, and geometric average. ’-‘ denotes for no ensembling, only base RPN is applied. The design option for current implementation is in bold.}
    \vspace{-5pt}
    \label{tab:rpn-abl}
\end{table}

\textbf{Bias-Balanced RPN.} To validate the effectiveness of our design, results on RPN recall and final detection precision for different classes of different RPN designs, including the ensembling strategy for both RPN outputs and the choice of unfrozen layers during finetuning, are evaluated in Table\ref{tab:rpn-abl}. Taking max as the ensembling strategy performs best, among other alternatives. Taking geometric average significantly degrades performance because any low objectness will produce a low final score. 
It can also be observed from the experiment that novel class AP is tightly related to AR of the RPN, while base class AP can remain stable with slightly inferior RPN AR, which validates one of our design philosophies to debias RPN thus improve novel class performance.
Unfreezing box regression layers and ensembling does not make much difference. Thus the extra computation overhead is not necessary.

\textbf{Re-detector.}
We study various design options in Re-detector, including the form of consistency loss (KL divergence, L1 difference, and negative cosine similarity between the normalized marginal probability distribution on base classes), layers in Re-detector to set unfrozen and classifier choice. As shown in Table\ref{tab:det-abl}, our current design maximizes the overall performance. Surprisingly, unfreezing more layers even lowers the performance. 

In addition, to validate the necessity of finetuning on both base and novel classes, we also implement a Re-detector where $f^n$ only detects $\mathcal{C}_n$. It produces relatively low results, probably due to severer false positives as diverse objects in base classes are encountered during test time, but unseen during finetuning, and the model is trained to categorize these objects the same as those deactivated background features from only a handful of samples, which is undoubtedly challenging.

\textbf{Inference time.} We report average inference time per image on COCO 2014 test set by adding modules into Faster R-CNN in Table\ref{tab:inference}. 
The inference time of Meta R-CNN\cite{yan2019meta} is also provided as a reference for representative meta-learning methods that demand exemplars for inference, which also introduces much lower extra computation than others, to the best of our knowledge.
As most weights are set frozen and shared, Retentive R-CNN introduces little overhead during test time to realize few-shot detection without forgetting, especially compared to meta learned models requiring exemplars at test time.

\begin{table}[t]
\small
    \centering
    \begin{tabular}{ c c c c |c c c}
        \Xhline{1pt}
          \textbf{$\mathcal{C}(f^n)$} & \textbf{$\mathcal{L}_{con}$} & \textbf{layers} & \textbf{cls} & \textbf{AP} & \textbf{bAP} & \textbf{nAP} \\\hline
        \textbf{all} &  \textbf{KLDiv} & \textbf{c+b} & \textbf{cos} & \textbf{32.1} & 39.2 & \textbf{10.5} \\
        all & L1 & c+b & cos & 32.0 & 39.2 & 10.3\\
        all & cos & c+b & cos & 31.9 & 39.2 & 10.0\\
        novel  & - & c+b & cos & 31.6 & 39.2 & 8.7\\
        all & KLDiv & c+b+h & cos & 31.9 & 39.2 & 9.8\\
        all & KLDiv & c+b & fc & 31.9 & \textbf{39.3} & 9.9\\\Xhline{1pt}
    \end{tabular}
    \caption{Ablation results in Re-detector design. $\mathcal{C}(f^n)$ represents the classification domain of a novel detector. The column of layers denotes unfrozen layers in the second stage, c represents classification, b represents bbox and h represents linear layers in the box head. The design option for current implementation is in bold.}
    \vspace{-5pt}
    \label{tab:det-abl}
\end{table}
\begin{table}[htb]
\small
    \centering
    \begin{tabular}{ c c c c c}
        \Xhline{1pt}
          \textbf{FRCN} & +BB-RPN & +ReDET & \textbf{ours} & \textbf{Meta R-CNN}\\\hline
         70.2 & 70.5 & 74.2 & 75.7 & 85.4 \\\Xhline{1pt}
    \end{tabular}
    \caption{Inference time in milliseconds on MS-COCO dataset. Meta R-CNN is reported in 10-shot from the original paper.}
    \label{tab:inference}
\end{table}

\begin{figure*}[ht]
\centering
\includegraphics[width=17.5cm]{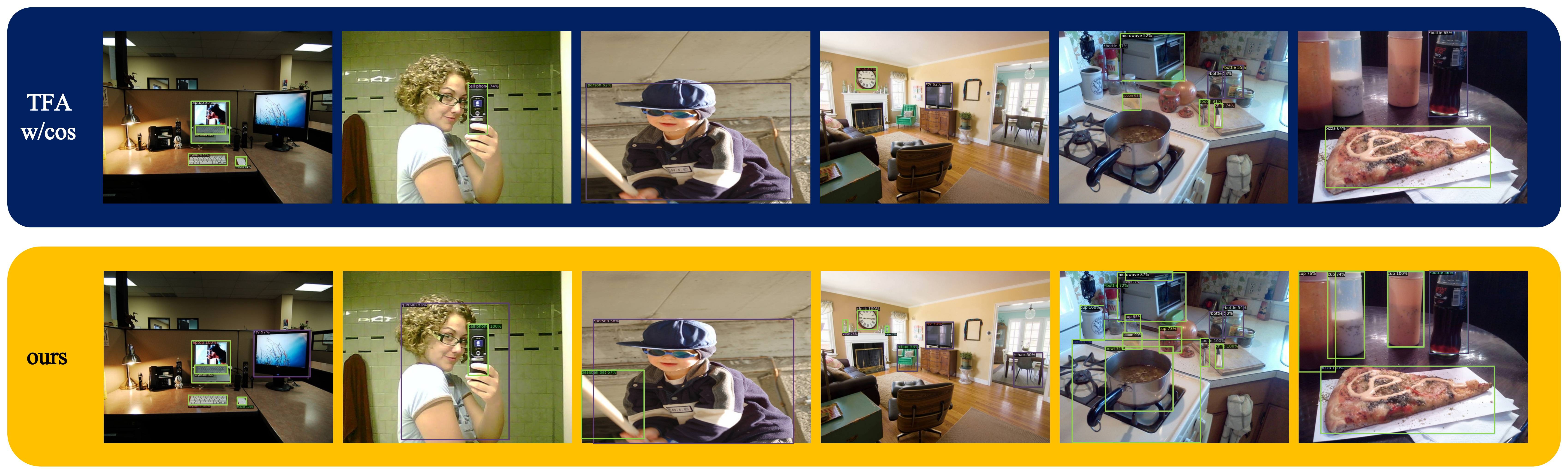}
\vspace{-1.5em}
\caption{Visualization of Retentive R-CNN and TFA w/cos\cite{wang2020frustratingly} results under MS-COCO 10-shot setting. Novel classes are bounded with purple boxes while base classes are bounded with green ones. Our method generally performs better on base classes and can detect novel class objects ignored by TFA in certain cases.}
\label{fig:vis}
\end{figure*}

\begin{figure*}[h]
\centering
\includegraphics[width=14.5cm]{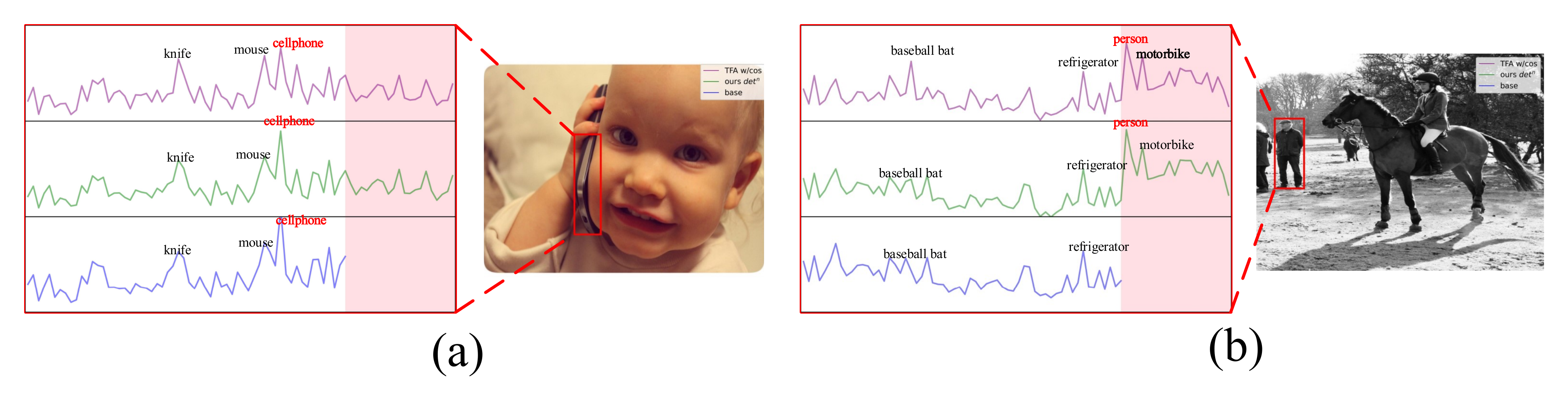}
\vspace{-1em}
\caption{Visualization of classification logits of Retentive R-CNN $det^n$ head, TFA w/cos\cite{wang2020frustratingly} (finetuned from the same model), and the base model's classifiers. Regions colored pink represent novel class logits. The consistency loss regularizes our model to produce similar distribution as the base model on base classes, making detection on (a) base classes more reliable and (b) novel classes less confusing.}
\vspace{-1em}
\label{fig:logit}
\end{figure*}

\textbf{Visualization.} We provide exemplary results obtained by Retentive R-CNN and TFA w/cos\cite{wang2020frustratingly} in Figure\ref{fig:vis} for comparison under MS-COCO 10-shot setting. The non-forgetting property of our method can be observed from the last four images containing either crowded scenes or less salient instances where TFA\cite{wang2020frustratingly} tends to ignore some of these objects, \textit{e.g.}, the inconspicuous baseball bat in the third image is ignored, and many well-learned objects are overseen in the fourth image by TFA\cite{wang2020frustratingly}. We also perform better on novel classes under certain cases, as shown in the first two images. 

We further investigate the role of consistency loss by comparing the classification distribution of our method and TFA w/cos\cite{wang2020frustratingly} and a base detector.
Specifically, we show two representative examples for the base class and novel class and visualize the logits of their classifiers for analysis. To make a fair comparison, both our method and TFA w/cos\cite{wang2020frustratingly} are trained upon this same base detector. 
It can be easily observed that our method produces much more similar logits distribution as the base model on base classes rather than TFA w/cos\cite{wang2020frustratingly}. 
Such property can better reserve base class performance, as shown in Figure\ref{fig:logit}(a), where the base model and ours produce unimodal distribution with one strong peak. When it comes to novel classes, as shown in Figure\ref{fig:logit}(b), base class distribution is suppressed, thus making a more confident response to novel classes.



\section{Conclusion}
In this paper, we have presented Retentive R-CNN to tackle the problem of G-FSD and proved that few-shot learning without forgetting is achievable in object detection. We analyze transfer learning based few-shot detection and find useful properties that are neglected by the community. Towards utilizing these properties, Retentive R-CNN is designed to combine base and novel detector simply and effectively, with Bias-Balanced RPN alleviating the bias of pretrained RPN and Re-detector reliably finding objects of both base and novel classes. Experiments on well-established few-shot object detection benchmarks show that Retentive R-CNN does not degrade on the base class while remains competitive on novel classes, reaching state-of-the-art overall performance among all data settings. Ablation study validates the effectiveness of our design. Nevertheless, the huge performance gap between few-shot and general object detection on data-limited classes indicates that this task is arduous by nature, and we hope that this paper sheds light on works to further boost novel class metrics with little or no trade-off on base classes.

\section*{Acknowledgement}
This research was partially supported by National Key R\&D Program of China (No. 2017YFA0700800), and Beijing Academy of Artiﬁcial Intelligence (BAAI).


\newpage
{\small
\bibliographystyle{ieee_fullname}
\bibliography{egbib}
}

\end{document}



\maketitle
\thispagestyle{empty}


\begin{figure*}[h]
\centering
\includegraphics[width=17.5cm]{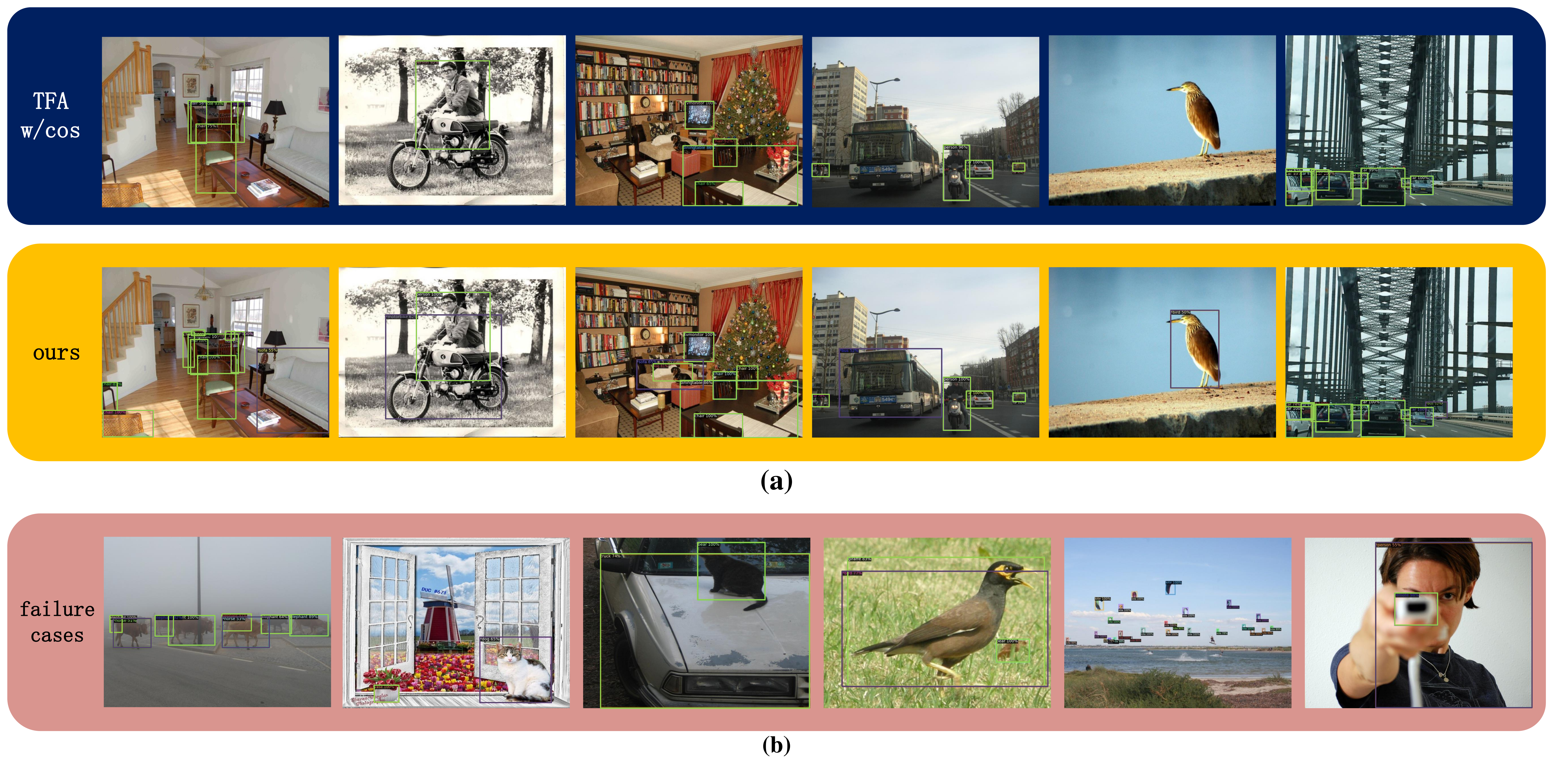}
\caption{(a) Visualization of Retentive R-CNN and TFA w/cos\cite{wang2020frustratingly} under Pascal VOC split1 2-shot settings. (b) Typical failure cases of Retentive R-CNN.}
\label{fig:vis}
\end{figure*}

\section{Implementation Details}
\textbf{Retentive R-CNN.} As a transfer learning based method, Retentive R-CNN is trained in two steps: the first step is trained on $\mathcal{D}^b$, which follows the same hyperparameters and learning schedule as in TFA\cite{wang2020frustratingly}; the second step is trained on a balanced dataset of both $\mathcal{C}^b \cup\mathcal{C}^n$. During the finetuning stage, we set learning rate to 0.05, coefficient for consistency loss to 0.1 across all settings, and only the finetuned RPN objectness is used. Note that the model is trained until full convergence; thus, the learning schedule for finetuning may vary from different datasplits. During inference, the base detector's classification logits are padded with 0 on the novel class entries; then, softmax operation is conducted on the padded logits to produce classification scores. As all activation in the network is ReLU and the base detector utilizes an fc classifier, logits with zero value can make good prior probabilities for novel classes, thus balance the scale of scores as the number of class entries are less than the novel detector. This improves base class AP and overall AP, \textit{e.g.}, overall AP increases from 32.0 to 32.1 under MS-COCO 10-shot setting. The novel detector also predicts base class probabilities, so we include these predictions for the non-maximum suppression procedure as well. Though consistency loss enhances the similarity between the prediction of the base detector and novel detector on base classes, the novel detector's base class predictions show ensembling effect to a certain extent, improving 0.05-0.1 base class AP upon base class AP of the pretrained model.

\textbf{Meta R-CNN\cite{yan2019meta} \& FsDetView\cite{kang2019few-shot}.} These two meta-learning methods are originally finetuned on randomly selected samples; we fix the samples to be the same as ours for fintuning for a fair comparison. Note that in both works, they finetune with base class samples as much as three times more than novel class samples to maintain base class performance, while we use the same number of samples to make a fair comparison. As Meta R-CNN\cite{yan2019meta} does not provide code for training on MS-COCO in the published implementation, we train Meta R-CNN\cite{yan2019meta} with identical hyperparameters and settings as FsDetView\cite{kang2019few-shot} on MS-COCO, which is implemented on the top of Meta R-CNN\cite{yan2019meta}.

\section{Examples for the Base Detector Rejecting Novel Class Instances}
Here we show more detection results from the pretrained base model in Figure\ref{fig:bg} to better demonstrate the property that the pretrained detector can reject novel class instances even if they are of great saliency to humans. The images are randomly selected from the first 100 images ordered by image id of MS-COCO 2014 minival set without cherry-picking. We bound the unrecognized novel class instances with black boxes and the detected objects with green boxes and their corresponding predicted category. Obviously, the base detector has a strong ability to ignore novel classes, thus false positives seldom occur from the base detector when encountering unseen classes. This property is utilized in Retentive R-CNN to maintain base class performance.

\section{More Detection Results \& Failure Case Analysis}
\begin{figure*}[htbp]
\centering
\includegraphics[width=17.5cm]{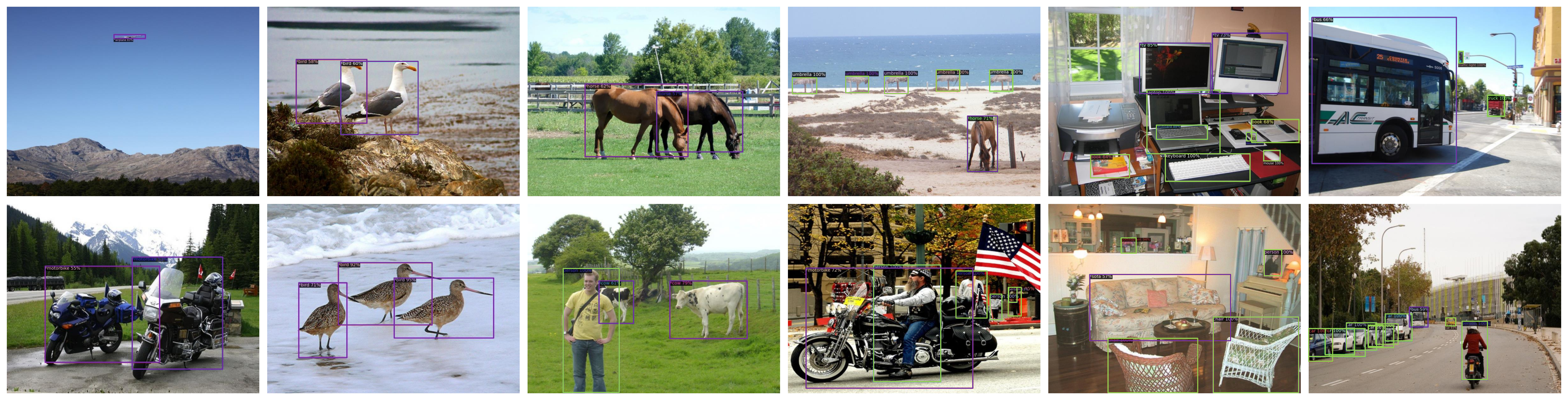}
\caption{Extra visualization of the detection results from Retentive R-CNN. The first row shows results under MS-COCO 10-shot settings while the second row shows results under Pascal VOC split1 2-shot settings.}
\label{fig:extra}
\end{figure*}

In this section, we show some extra detection results for further demonstration of the effectiveness of our method and a qualitative failure case analysis. Figure\ref{fig:vis}(a) shows representative results for comparing our method and TFA w/cos\cite{wang2020frustratingly} under Pascal VOC split1 2-shot setting. The conclusion is consistent with the qualitative comparison shown in the main paper that our method typically performs better on base classes due to the non-forgetting property and reduces object confusion on novel class instances, successfully detecting many of the ignored objects by TFA w/cos\cite{wang2020frustratingly}. Some extra detection visualization of our method is shown in Figure\ref{fig:extra}.

Nevertheless, both our method and previous works have a vast metrics gap between few-shot classes and classes trained from abundant data, indicating that few-shot object detection is still hard by nature. We analyze several typical failure patterns in Figure\ref{fig:vis}(b). The first four columns show false positive cases, mainly due to: 1) Though not common, the base detector sometimes produces false positives on unseen objects, producing overlapped predicted boxes of both base and novel categories on the same instance; 2) features are not discriminative enough for few-shot categories, thus confusion among classes like misclassification among few-shot classes and domination of base classes over novel classes. The fifth column shows a typical case for transfer learning based methods where novel class objects are hard to be detected due to deactivation in the backbone, showing that such bias caused by pretraining is hard to be alleviated. The sixth column shows another failure pattern caused by box regression, probably because accurate localization for categories with complex shapes is also challenging to learn under low-shot scenarios.

\section{Results over Multiple Runs}
To show the effectiveness of our method without random effect, we ran our model over 10 sets of random samples under 5, 10, 30-shot settings on COCO dataset, using exactly the same samples as TFA\cite{wang2020frustratingly}. The results are shown in Table\ref{tab:random}. We obtain better performance in terms of all metrics (AP, bAP, nAP) under each of these settings.

\begin{table*}
    \centering
    \begin{tabular}{c | c c c | c c c | c c c}
      \hline
      \multirow{2}{*}{\textbf{Methods}} &
      \multicolumn{3}{c |}{\textbf{5 shot}}  &
      \multicolumn{3}{c |}{\textbf{10 shot}}  &
      \multicolumn{3}{c}{\textbf{30 shot}}\\
   & \textbf{AP} &  \textbf{bAP} & \textbf{nAP} & \textbf{AP} &  \textbf{bAP} & \textbf{nAP} & \textbf{AP} &  \textbf{bAP} & \textbf{nAP} \\\hline
      Retentive R-CNN & \textbf{31.4} & \textbf{39.3} & \textbf{7.7} & \textbf{31.8} & \textbf{39.2} & \textbf{9.5} & \textbf{32.6} & \textbf{39.3} & \textbf{12.4} \\\hline
      FRCN-ft-full\cite{wang2020frustratingly} & 14.4 & 17.6 & 4.6 & 13.4 & 16.1 & 5.5 & 13.5 & 15.6 & 7.4 \\
      TFA w/ fc\cite{wang2020frustratingly} & 25.6 & 31.8 & 6.9 & 26.2 & 32.0 & 9.1 & 28.4 & 33.8 & 12.0 \\
      TFA w/ cos\cite{wang2020frustratingly} & 25.9 & 32.3 & 7.0 & 26.6 & 32.4 & 9.1 & 28.7 & 34.2 & 12.1 \\\hline
    \end{tabular}
    \caption{Results over \textbf{10 random runs} on COCO dataset under 5, 10, 30-shot settings. Note that we use the same samples as TFA\cite{wang2020frustratingly} so that the metrics are directly comparable. We obtain better performance in terms of all metrics.}
    \label{tab:random}
\end{table*}

\newpage
{\small
\bibliographystyle{ieee_fullname}
\bibliography{egbib}
}